\documentclass[journal,twoside,web]{IEEEcolor}
\usepackage{generic}
\usepackage{cite}
\usepackage{amsmath,amssymb,amsfonts}
\usepackage{booktabs}
\usepackage{graphicx}
\usepackage{xcolor}
\usepackage{url}
\usepackage{tikz}
\usepackage{pgfplots}
\usepackage{textcomp}
\usepackage{marvosym} 
\usepackage{subcaption}
\usepackage{multirow}

\makeatletter
\let\NAT@parse\undefined %
\makeatother

\usepackage{hyperref}
\usepackage{orcidlink}

\IEEEaftertitletext{
\begin{center}
\vspace{-3em}
\footnotesize
\textit{This work has been submitted to the IEEE for possible publication. Copyright may be transferred without notice, after which this version may no longer be accessible.}
\end{center}
}

\def\BibTeX{{\rm B\kern-.05em{\sc i\kern-.025em b}\kern-.08em
    T\kern-.1667em\lower.7ex\hbox{E}\kern-.125emX}}

\definecolor{revisionblue}{RGB}{0,0,180}

\providecommand{\refname}{References}


\begin{document}

\title{Spiking Neural Networks for fMRI-Based Visual Semantic Decoding}

\author{%
\makebox[\textwidth][c]{%
\begin{minipage}{\textwidth}
\centering
Jiahong~Zhang\orcidlink{0000-0002-5687-1839},
Jinning~Zhao\orcidlink{0009-0006-7287-3953},
Sijun~Shen\orcidlink{0009-0009-0654-6358},
Siyuan~Xu\orcidlink{0009-0004-4169-3488},
Bo~Xu\orcidlink{0000-0002-1111-1529},
and~Guoqi~Li\orcidlink{0000-0002-8994-431X}
\end{minipage}%
}%
\thanks{Jiahong Zhang and Jinning Zhao contributed equally to this work.}%
\thanks{Jiahong Zhang, Jinning Zhao, Siyuan Xu, Bo Xu, and Guoqi Li are with the Institute of Automation, Chinese Academy of Sciences, Beijing 100045, China, and with the School of Artificial Intelligence, University of Chinese Academy of Sciences, Beijing 100049, China.}%
\thanks{Sijun Shen is with the State Key Laboratory of Media Convergence and Communication, Communication University of China, Beijing 100024, China.}%
\thanks{Corresponding author: Guoqi Li.}}

\maketitle

\begin{abstract}
Functional magnetic resonance imaging (fMRI)-based visual decoding aims to recover visual information from measured brain activity, commonly by mapping fMRI responses into latent visual features for downstream decoding tasks. 
Most existing methods learn mappings from fMRI responses to visual features extracted by artificial neural networks (ANNs), yet it remains unclear whether ANN-derived features provide suitable targets for brain decoding.
In this study, we investigate spiking neural network (SNN)-derived visual features as alternative targets for fMRI-based visual decoding. 
We compare an ANN baseline with four SNN variants from the same architectural family, which differ in their spiking dynamics. 
To isolate the effect of the target features, all models use the same L2-regularized linear fMRI-to-feature decoder, while only the feature vectors used as regression targets are varied. 
Compared with the ANN baseline, SNN-derived features exhibit stronger alignment with fMRI responses and improve visual semantic decoding performance. 
For instance, on the GoD dataset, SNN-derived features reduce feature-prediction error from 0.7707 to 0.0282 and improve top-1 semantic decoding accuracy from 0.1800 to 0.4400. 
Ablation results further indicate that both spiking neural dynamics and temporal simulation steps contribute to the observed advantage. 
These findings support SNN-derived features as effective brain-decodable visual representations and highlight target feature design as an important component of fMRI-based visual decoding.
\end{abstract}


\section{Introduction}
\label{sec:intro}

{Functional magnetic resonance imaging (fMRI) is a major noninvasive tool for investigating visual representations in the human brain}~\cite{logothetis2008we}. In fMRI-based visual decoding, the goal is to infer the viewed stimulus from measured brain activity. This capability provides a principled means to probe cortical visual representations and advance translational brain-computer interface technologies~\cite{kay2008identifying,nishimoto2011reconstructing,zafar2015decoding,yin2026cogformer}. 

Despite substantial progress, visual decoding from fMRI remains a challenging inverse problem~\cite{naselaris2009bayesian,ozcelik2023natural}. Natural images are characterized by rich hierarchical structure, spanning low-level appearance, object-level attributes, and high-level semantic context. In contrast, fMRI signals provide an indirect, delayed, and spatially distributed measurement of neural activity through blood-oxygen-level-dependent (BOLD) responses~\cite{logothetis2008we}. This discrepancy creates a fundamental representational gap between visual experience and measurable brain activity~\cite{fang2023alleviating,ferrante2024retrieving}. Consequently, effective decoding cannot rely solely on direct mappings from voxels to labels or pixels. Instead, it requires an intermediate representation space in which neural responses and visual semantics can be reliably aligned.

Modern visual decoding pipelines commonly address this issue by mapping brain activity into visual features learned by pretrained visual models for image reconstruction~\cite{shen2019deep,seeliger2018generative,takagi2023high}, or into vision-language representations for semantic classification and image retrieval~\cite{radford2021learning,ferrante2024retrieving}. This strategy has produced strong empirical results and has become a popular design pattern in fMRI-based visual decoding. 

Within this framework, the choice of target {visual features} remains an important methodological consideration. In most recent studies, these features are inherited from artificial neural networks (ANNs), such as convolutional classifiers, multimodal embedding models, or generative backbones~\cite{wang2020neural,zhou2022exploring,jing2025beyond,dai2025mindaligner}. Although these ANN-derived features are powerful and readily integrated into existing pipelines, they are typically adopted for their success in computer vision rather than for their demonstrated alignment with human brain representations. Consequently, many decoding methods focus on learning an accurate mapping from fMRI signals to chosen ANN features, while the suitability of the target features themselves receives less attention~\cite{xia2025exploring,huang2025seeing}. This issue matters because the predicted features constrain what information the decoder can recover, how errors affect downstream tasks, and how the resulting brain-to-vision mapping should be interpreted~\cite{cammarota2026bridging}.

To this end, we consider whether the brain-inspired visual features could provide a more decodable target for fMRI-based visual decoding. Spiking neural networks (SNNs) are a natural candidate, as they encode information through brain-inspired discrete spike events and explicit temporal neural dynamics~\cite{maass1997networks,tavanaei2019deep,stbp}. Recent advances further show that deep SNNs can achieve competitive performance on visual recognition tasks~\cite{stbp,wu2019direct,mpsn,sew,shen2026stage}, indicating that spike-based models can provide effective visual representations. Nevertheless, there is still little direct evidence on whether SNN-derived visual features are more suitable than ANN-derived visual features for fMRI-based visual decoding.

In this work, we address this question through a controlled comparison of ANN- and SNN-derived visual features for fMRI-based visual decoding. We use the same ridge-regression decoder for all models. The decoder maps fMRI activity to visual features extracted from either a standard ResNet-18 ANN backbone~\cite{ResNet18} or an SNN variant from the same architectural family.
We consider four SNN variants: leaky integrate-and-fire (LIF)~\cite{lif}, parallel spiking neuron (PSN)~\cite{psn}, memory-based parallel spiking neuron (MPSN)~\cite{mpsn}, and bursting spiking neuron (BuSNN)~\cite{zhang2026burst}. {These variants cover different forms of spiking dynamics. LIF represents classical membrane integration. PSN introduces parallel temporal computation. MPSN further adds explicit memory propagation. BuSNN models burst-like spike coding. This progression allows us to examine which neural dynamics contribute to the decoding differences.}
By keeping the decoder and evaluation protocol fixed and varying only the {predicted visual features}, this design allows us to isolate the role of the visual features themselves.

We evaluate the resulting visual features on public fMRI benchmarks using a multi-level protocol that includes voxel-level alignment, semantic-level alignment, semantic-based classification, retrieval, and reconstruction. On Mini-Algonauts 2021~\cite{cichy2021algonauts}, voxel-level analyses show that SNN representations better correspond to measured brain responses than ANN representations. Across Generic Object Decoding (GoD)~\cite{god} and Natural Scenes Dataset (NSD)~\cite{nsd}, SNN-derived visual features are more accurately predicted from fMRI activity and lead to stronger semantic decoding than ANN-derived features. Together, these results suggest that SNN-derived visual features are promising targets for fMRI-based decoding and highlight visual-feature choice as an important scientific variable rather than a fixed engineering choice.

The main contributions of this study are as follows.
\begin{enumerate}
\item We introduce a new perspective for fMRI-based visual decoding by treating the target visual representation as a key scientific variable. Instead of only improving the brain decoder, we examine whether SNN-derived visual features provide a more fMRI-predictable alternative to conventional ANN-derived features.

\item We develop a controlled ANN--SNN comparison framework to isolate the effect of target feature choice. Using the same L2-regularized fMRI-to-feature decoder, we compare a ResNet-18 ANN with multiple SNN variants from the same architectural family, thereby enabling a direct and fair evaluation of spike-based visual features.

\item We show empirical evidence that SNN-derived features improve fMRI decoding across multiple levels. Further analyses of SNN variants and time steps show that spiking neural dynamics contribute to these gains, supporting SNN-derived features as effective brain-decodable 
targets for visual decoding.
\end{enumerate}

\begin{figure*}[!t]
\centering
\includegraphics[width=\textwidth]{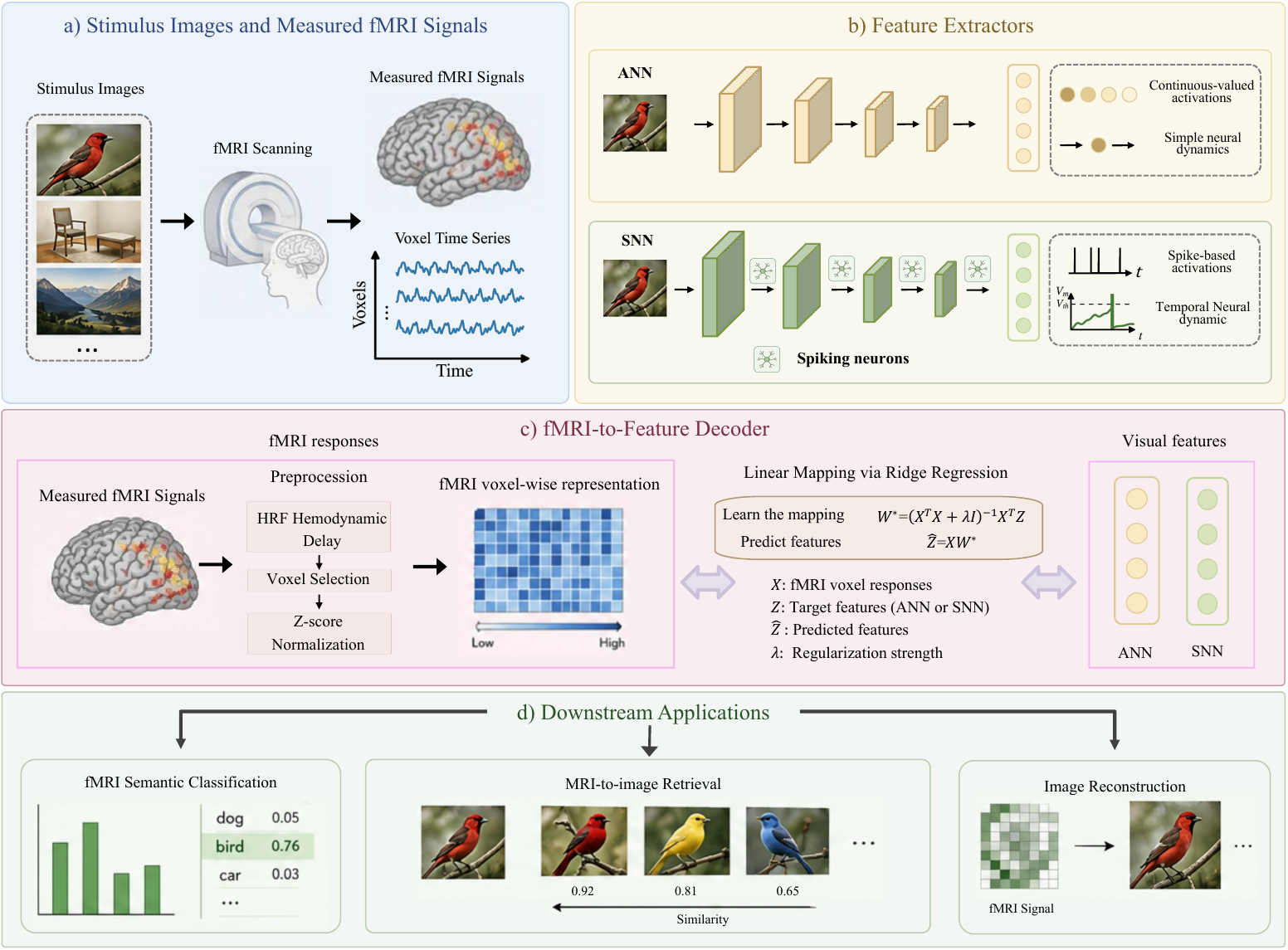}
\caption{Evaluation pipeline for comparing ANN and SNN visual features in fMRI-based visual semantic decoding. a) Stimulus images are presented to subjects during fMRI scanning, producing voxel-wise BOLD responses. b) The same images are processed by either an ANN feature extractor or an SNN feature extractor, where ANN features are {dense} and static, while SNN features are spike-based and temporally dynamic. c) Using the same fMRI-to-feature decoder, standardized fMRI responses are mapped into the corresponding ANN or SNN model-derived features. d) The predicted fMRI-decoded features are then evaluated in downstream semantic classification, image retrieval, and image reconstruction tasks. Since the decoding pipeline is fixed across models, performance differences mainly reflect the advantage of SNN features for fMRI decoding.}
\label{fig:overview}
\end{figure*}

\section{Related Work}
\label{sec:related}

\textbf{Spiking neural networks.}
Spiking neural networks (SNNs) represent information through discrete spike events, spike timing, and membrane dynamics~\cite{maass1997networks,lif}. This makes them a brain-inspired alternative to conventional ANNs with continuous activations. Early SNN training was difficult because spike generation is non-differentiable. This issue has been largely alleviated by surrogate-gradient learning and modern SNN training frameworks~\cite{neftci2019surrogate,eshraghian2023training,tavanaei2019deep,roy2019towards}. Recent deep SNNs have achieved competitive visual recognition performance~\cite{stbp,wu2019direct}. Residual SNNs and advanced neuron variants further improve network depth, temporal computation, and representation capacity~\cite{sew,psn,mpsn}. However, most SNN studies focus on classification accuracy, training efficiency, or event-driven inference. It remains less explored whether spike-based visual representations are easier to infer from brain measurements or better aligned with fMRI responses. This work addresses this question by evaluating SNN-derived visual features as decoding targets under a controlled fMRI-to-feature mapping framework.

\textbf{fMRI visual semantic decoding.}
fMRI visual decoding has progressed from category-level pattern analysis to feature-based semantic decoding and image reconstruction~\cite{haxby2001distributed,kay2008identifying,miyawaki2008visual,naselaris2009bayesian,nishimoto2011reconstructing,horikawa2017generic,shen2019deep}. Recent studies further map fMRI responses into vision-language or generative feature spaces~\cite{du2023decoding,liu2023brainclip,takagi2023high,ozcelik2023latent,mindeye2023,dream2024}. These feature spaces support semantic classification, image retrieval, captioning, and natural-scene reconstruction. Together, these studies show that the target representation strongly affects what visual information can be recovered from fMRI activity. However, most existing pipelines use ANN-derived visual features, multimodal embeddings, or generative features as default targets. In contrast, our study investigates whether SNN-derived spike-based representations can provide more brain-decodable visual features for fMRI-based visual semantic decoding. 


\section{Materials and Methods}
\label{sec:methods}

\subsection{Problem Formulation and Framework Overview}

Let {$I_i \in \mathcal{I}$} denote the {stimulus image} presented in the $i$-th trial, and let
$x_i \in \mathbb{R}^{{p}}$ denote the corresponding preprocessed fMRI voxel response from
{$p$} selected voxels. Given a set of paired samples:
\begin{equation}
\mathcal{D}=\{(x_i,{I_i})\}_{i=1}^{N},
\end{equation}
the goal of fMRI-based visual semantic decoding is to infer the visual content of {$I_i$} from
the measured brain response $x_i$.

This decoding problem is challenging because visual stimuli contain rich semantic information, whereas fMRI responses provide noisy and indirect voxel-wise measurements of brain activity. Following recent fMRI retrieval and reconstruction studies~\cite{ferrante2024retrieving}, we therefore formulate decoding in terms of {visual features} rather than pixels or labels.

Let $m \in \mathcal{M}$ index a pretrained visual encoder, where
$\mathcal{M}$ includes the ANN baseline and the SNN variants considered in this study.
Each encoder maps the same stimulus image to model-derived {features}:
\begin{equation}
z_i^{(m)} = f_m({I_i}), \qquad z_i^{(m)} \in \mathbb{R}^{d_m},
\label{eq2}
\end{equation}
where $f_m(\cdot)$ denotes the feature extractor of encoder $m$ after removing the final classification head, and $d_m$ is the dimensionality of its feature space.

For a given encoder $m$, the decoding problem is to estimate from $x_i$ an
fMRI-decoded feature $\hat{z}_i^{(m)}$ that is consistent with the model-derived feature $z_i^{(m)}$:
\begin{equation}
\hat{z}_i^{(m)} \approx z_i^{(m)}.
\end{equation}

As shown in Fig.~\ref{fig:overview}, the overall framework is organized as a controlled comparison of candidate visual feature spaces. For each visual encoder, target features are first extracted from the stimulus images. ANN features are obtained from a conventional ResNet-18 backbone~\cite{ResNet18}, while SNN features are obtained from spiking variants with the same backbone family. The preprocessed fMRI responses are then mapped to each feature space using the same regularized linear fMRI-to-feature decoder. For each ANN or SNN feature space, a separate mapping is trained, but the decoder form, input preprocessing, and training protocol are kept fixed.

During testing, an unseen fMRI response is projected into the selected ANN or SNN feature space. The decoded feature \(\hat{z}_i^{(m)}\) is then evaluated by semantic classification, fMRI-to-image retrieval, and semantic-guided reconstruction. Under this formulation, performance differences mainly reflect the choice of target visual representation rather than differences in decoder complexity. The central question of this study is whether SNN-derived visual features provide a more suitable decoding target than ANN-derived features for fMRI-based visual semantic decoding.

\subsection{ANN and SNN Visual Features}

To compare conventional and spike-based visual features under the same
fMRI decoding framework, we extracted features from an ANN backbone and
from several SNN backbones. All visual encoders were pretrained on ImageNet~\cite{imagenet} and kept fixed during fMRI decoder training. The extracted features were used only as {candidate visual features} for fMRI-to-feature decoding.

\subsubsection{ANN representation}
The ANN baseline is a standard ResNet-18 model~\cite{ResNet18}. Given a stimulus image {$I$}, the
network computes a sequence of continuous-valued activations through convolutional
layers, normalization layers, and nonlinear transformations. We remove the final
classification layer and use the penultimate feature as the ANN {visual feature target}:
\begin{equation}
z_{\mathrm{ANN}} = f_{\mathrm{ANN}}({I}).
\end{equation}
This representation is obtained from a single feed-forward pass and consists of dense
continuous activations. It therefore provides the conventional {visual features} against which
the SNN-derived features are compared.

\subsubsection{SNN representation}

For the SNN models, we use SEW-ResNet-18~\cite{sew}, which preserves the residual block structure and layer configuration of ResNet-18 while replacing conventional activation units with spiking neurons and spike-based residual operations. Therefore, the ANN and SNN backbones are architecturally comparable, and their main difference lies in neural dynamics used to generate model-derived features.

Unlike ANN activations obtained from a single feed-forward pass, SNN neurons process a visual stimulus over \(T\) discrete time steps and generate spike responses through internal state updates. For a spiking neuron at time step \(t\), we denote the input current, hidden state, membrane potential, and output spike as \(X_t\), \(H_t\), \(V_t\), and \(S_t\), respectively. The general spike-generation process can be summarized as:
\begin{equation}
\begin{aligned}
H_t &= \phi(V_{t-1}, X_t), \\
S_t &= \Theta(H_t - V_{\mathrm{th}}), \\
V_t &= \psi(H_t, S_t),
\end{aligned}
\end{equation}
where \(\phi(\cdot)\) denotes the charging function, \(\psi(\cdot)\) denotes the reset function, \(V_{\mathrm{th}}\) is the firing threshold, and \(\Theta(\cdot)\) is the Heaviside step function. Different SNN variants correspond to different choices of these spiking dynamics. 

For the same input image \(I\), the SNN produces time-dependent feature responses $\{f_{\mathrm{SNN}}^{t}(I)\}_{t=1}^{T}$, where \(f_{\mathrm{SNN}}^{t}(I)\) is the feature vector extracted at time step \(t\) after removing the final classification head. To obtain a fixed-length representation for fMRI decoding, we aggregate these features by temporal averaging:
\begin{equation}
z_{\mathrm{SNN}} = \frac{1}{T}
\sum_{t=1}^{T}
f_{\mathrm{SNN}}^{t}(I).
\end{equation}
Thus, the SNN feature encodes visual information through temporally accumulated spike-based responses rather than a single dense activation vector.

\subsubsection{Spiking neuron variants}

We evaluate four SNN variants with different spiking dynamics under the same SEW-ResNet-18~\cite{sew} backbone family.
All variants receive time-indexed input currents $\{X_t\}_{t=1}^{T}$ and generate output spikes $\{S_t\}_{t=1}^{T}$, but they differ in how the state variables ${H_t,V_t}$ and memory state $M$ are used to transform temporal inputs into spike responses.

LIF~\cite{lif} represents the classical sequential spiking mechanism.
It updates the charging, firing, and reset states step by step across time and generates spikes through threshold-based firing.
PSN~\cite{psn} introduces a parallel spiking formulation, where multi-step inputs are transformed into time-indexed state variables and spike outputs with reduced strict sequential dependency.
MPSN~\cite{mpsn} further incorporates a memory state $M$ to strengthen temporal information propagation across simulation steps.
BuSNN~\cite{zhang2026burst} uses a memory-guided burst mechanism, where burst-like spike responses provide a more selective event-driven representation. 


By keeping the backbone and decoding pipeline fixed while varying the spike-generation mechanism, this design allows us to examine whether the decoding advantage is tied to a specific neuron model or reflects a broader property of SNN-derived visual features.

\subsection{{Ridge Regression fMRI-to-feature Decoder}}

For each visual encoder $m \in \mathcal{M}$, where $\mathcal{M}$ includes the ANN baseline and the SNN variants, we train an independent fMRI-to-feature decoder to map preprocessed voxel responses to model-specific visual features. Let $\mathcal{D}_{\mathrm{tr}}$ denote the training split. For a training pair $(x_i,I_i)\in \mathcal{D}_{\mathrm{tr}}$, let $x_i \in \mathbb{R}^{p}$ be the preprocessed fMRI voxel response.

The fMRI-to-feature decoder for encoder $m$ is defined as a linear ridge-regression mapping:
\begin{equation}
\hat{z}_i^{(m)} = g_m(x_i;W_m)=x_i^{\top}W_m ,
\end{equation}
where $W_m \in \mathbb{R}^{p \times d_m}$ denotes the learned mapping from the voxel space to the feature space of encoder $m$.

The mapping is fitted by solving an L2-regularized least-squares problem:
\begin{equation}
W_m^{*}
=
\arg\min_{W_m}
\sum_{(x_i,I_i)\in \mathcal{D}_{\mathrm{tr}}}
\left|
x_i^{\top}W_m - z_i^{(m)}
\right|_2^2
+
\lambda |W_m|_F^2 .
\end{equation}
Here, $\lambda$ controls the L2 regularization strength.

After fitting the ridge-regression model on the training split, the optimized mapping $W_m^{*}$ is applied to unseen fMRI responses. For a test sample $x_j$, the decoded feature is:
\begin{equation}
\hat{z}_j^{(m)} = x_j^{\top} W_m^{*}.
\end{equation}

This training procedure is repeated independently for each ANN or SNN target feature set. Across different choices of $m$, the visual encoder and target feature space change, whereas the decoder form, input preprocessing, regularization strength, and training protocol are kept fixed. Therefore, performance differences mainly reflect the choice of predicted visual features rather than additional decoder flexibility.

\begin{figure*}[!t]
\centering
\includegraphics[width=\textwidth]{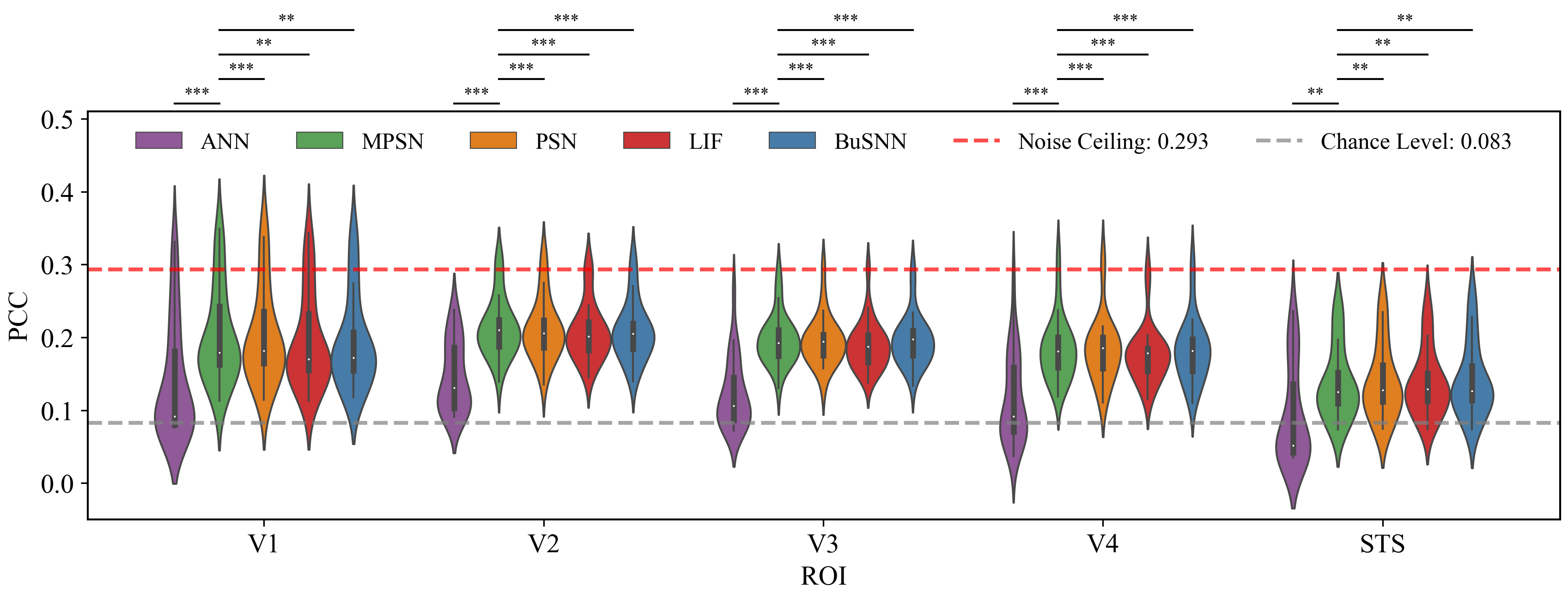}
\caption{{Voxel-level alignment between model-derived representations and measured fMRI responses on Mini-Algonauts 2021. PCC distributions are shown for ResNet18 and four SNN-based variants across V1, V2, V3, V4, and STS. Higher PCC indicates stronger alignment with voxel-wise fMRI responses. Statistical annotations denote one-tailed comparisons between each SNN variant and the ResNet18 baseline. Statistical significance was assessed using a one-tailed Welch's $t$-test
with the alternative hypothesis {that each SNN variant has a higher PCC than the ANN baseline}.
Brackets indicate comparisons between the baseline and each {SNN} variant.
Significance levels are denoted as ns, not significant; $^{*}p < 0.05$,
$^{**}p < 0.01$, and $^{***}p < 0.001$. Dashed lines indicate the empirical noise ceiling and a statistical chance level.}}
\label{fig:pcc}
\end{figure*}

\subsection{{Semantic Classification, Retrieval, and Reconstruction}}

At inference, an unseen fMRI response is mapped to a decoded feature $\hat{z}$ in the
selected ANN or SNN {feature representation}. The decoded feature is then used for downstream tasks.

For semantic decoding, $\hat{z}$ is passed to the classifier head of the corresponding visual backbone. This produces a category prediction from the fMRI-decoded feature. The same classifier head is also applied to the model-derived feature of the stimulus, so that semantic predictions from fMRI-decoded and model-derived features can be compared.

For fMRI-to-image retrieval, $\hat{z}$ is used as a query feature. Candidate stimulus images are first represented by the same ANN or SNN backbone. The candidate images are then ranked according to their feature similarity to $\hat{z}$, and the top-ranked
images are taken as the retrieved results.

For reconstruction, we follow the semantic-guided generation strategy used in recent fMRI decoding work~\cite{ferrante2024retrieving}. Specifically, the decoded feature is passed through the classifier head to obtain a predicted semantic label. This label is then converted into a text prompt and used to condition a pretrained text-to-image diffusion model~\cite{ho2020denoising}. The image generator is kept fixed, so the reconstruction differences mainly arise from the semantic labels decoded from ANN- or SNN-derived features. Although this procedure does not perform direct pixel-level reconstruction from fMRI, it provides a qualitative and semantic-level test of whether the decoded feature retains sufficient semantic information to guide image generation toward the perceived stimulus category.

\section{Experiments and Results}
\label{sec:results}

\subsection{{Datasets}}

{
We evaluated the proposed framework on three public fMRI datasets with complementary roles. Mini-Algonauts 2021~\cite{cichy2021algonauts} was used for voxel-level representation alignment, while GoD~\cite{god} and NSD~\cite{nsd} were used for semantic-level representation alignment and downstream semantic decoding. The GOD dataset contains fMRI data collected from five subjects during an image-presentation task. All visual stimuli were sourced from the ImageNet database, covering 150 training categories and 50 distinct test categories. For the NSD dataset, we used the data from Subject 1 and selected the NSD general ROI for the fMRI feature space. 
We randomly selected 10\% of the training samples from each dataset as a validation set for model selection while keeping the original test set unchanged. 
ANN and SNN features were compared using the same fMRI-to-feature decoder and fixed evaluation metrics, ensuring that performance differences were primarily attributable to the choice of visual features predicted from fMRI.
}


\subsection{Experimental Settings}
All experiments were implemented in Python 3.9 using PyTorch and scikit-learn, with SNN models built on the SpikingJelly framework~\cite{fang2023spikingjelly}. The visual encoders were pretrained on ImageNet~\cite{imagenet} and frozen during fMRI decoder training.
Images were resized to $224 \times 224$, and features were extracted from the penultimate layer with a fixed dimension of 512. The ANN baseline used ResNet-18~\cite{ResNet18}, while the SNN models used SEW-ResNet-18~\cite{sew} with different spiking neuron dynamics. For SNNs, the default simulation step was set to $T=4$, and time-dependent features were averaged across steps to obtain fixed-length representations.

For all features, we used the same data splits, fMRI inputs, ridge-regression decoder, and evaluation metrics. The decoder mapped fMRI responses to model-derived features with a fixed regularization coefficient $\alpha=1000$. 

\subsection{{Representation Alignment with fMRI Responses}}

\subsubsection{{Voxel-level alignment}}

To assess voxel-level brain alignment, we quantified the correspondence between model-derived visual representations and voxel-wise fMRI responses. The PCC between predicted and measured voxel responses was used as the evaluation metric and averaged within each ROI. Model selection was based on validation performance. 

We report PCC results for selected visual and semantic-related ROIs in Mini-Algonauts 2021, including V1, V2, V3, V4, and STS. Fig.~\ref{fig:pcc} summarizes the PCC distributions across these ROIs. We compare ResNet18 as the ANN baseline with four SNN-based variants, including MPSN~\cite{mpsn}, PSN~\cite{psn}, LIF~\cite{lif}, and BuSNN~\cite{zhang2026burst}.
Overall, the SNN-based representations show stronger voxel-level correspondence with fMRI responses than the ANN baseline. This trend is reflected by the higher PCC distributions of the SNN variants across the selected ROIs. 

This voxel-level analysis provides an upstream assessment of brain-response alignment before downstream decoding evaluation. {Visual features that better account for measured cortical responses may provide a more reliable decoding target for subsequent semantic decoding.} The PCC results therefore support the use of SNN representations as effective brain-decodable {visual features} for fMRI-based visual decoding.

\subsubsection{{Semantic-level alignment}}

We next examined whether fMRI responses can be aligned with the semantic representation spaces of ANN and SNN models. For each visual backbone, we used the pre-classification feature as its model-specific semantic representation, since this feature directly supports the final category prediction and therefore captures high-level visual semantics. Given an image stimulus, the visual model maps it to model-derived features, while the fMRI-to-feature decoder maps the paired fMRI response into the same feature space. Semantic-level alignment was measured by the mean squared error (MSE) between the fMRI-decoded features and the corresponding model-derived features. A lower MSE indicates stronger consistency between the semantic structure learned by the visual model and the representational structure encoded in fMRI activity.

Table~\ref{tab:combined_alignment} reports the MSE between fMRI-decoded features and model-derived features on GoD and NSD. Across both datasets, SNN-based semantic representations show substantially lower prediction errors than the ANN baseline. On GoD, the test MSE decreases from 0.7707 for ANN to 0.0282 for the best-performing SNN models. On NSD, the LIF model achieves the lowest test MSE, reducing the error from 0.7304 to 0.0123. PSN, MPSN, and BuSNN also outperform the ANN baseline, indicating that the improvement is not restricted to a single spiking neuron implementation.

We further report Mean Feature Correlation (MFC) in Table~\ref{tab:combined_alignment}, which measures the average Pearson correlation between predicted and target features across valid feature dimensions. Unlike MSE, MFC is less affected by the numerical scale of feature values and focuses on whether the feature-wise variation trends are preserved. 
Specifically, for each feature dimension $k$, we calculate the correlation coefficient between the predicted value $\hat{z}_k$ and true value $z_k$ on the test samples:
\begin{equation}
\rho_k = \frac{\mathrm{Cov}(\hat{z}_k, z_k)}{\sqrt{\mathrm{Var}(\hat{z}_k) \cdot \mathrm{Var}(z_k)}},
\end{equation}
Invalid dimensions with a standard deviation of less than $10^{-8}$ are excluded to ensure numerical stability. The final MFC value is the arithmetic mean of the correlation coefficients for all valid dimensions.

As shown in Table~\ref{tab:combined_alignment}, several SNN variants achieve higher test MFC than the ANN baseline. On GoD, PSN obtains the highest test MFC of 0.3116, followed by BuSNN and LIF. On NSD, BuSNN and PSN outperform ANN, while LIF and MPSN show lower MFC despite their lower MSE. These results suggest that SNN-derived representations generally reduce absolute prediction error, and some spiking variants further improve feature-wise semantic consistency.

These results indicate that SNN-derived semantic representations are more readily recoverable from fMRI responses under the same decoding model.

\begin{table}[!t]
\centering
\caption{Semantic-level alignment comparison between MSE (lower is better) and MFC (higher is better) metrics for fMRI-decoded features.}
\label{tab:combined_alignment}
\setlength{\tabcolsep}{2pt}
\begin{tabular}{llcccccc}
\toprule
Dataset & Model & \multicolumn{3}{c}{MSE} & \multicolumn{3}{c}{MFC} \\
\cmidrule(lr){3-5} \cmidrule(lr){6-8}
 & & Train & Val & Test & Train & Val & Test \\
\midrule
GoD & ANN~\cite{ResNet18} & 0.13948 & 0.93780 & 0.77072 & 0.95541 & 0.14782 & 0.26238 \\
& LIF~\cite{lif} & \textbf{0.00195} & \textbf{0.02329} & \textbf{0.02824} & \textbf{0.95555} & 0.12729 & 0.29203 \\
& PSN~\cite{psn} & 0.01163 & 0.07812 & 0.06354 & 0.95551 & \textbf{0.18922} & \textbf{0.31164} \\
& MPSN~\cite{mpsn} & 0.00530 & 0.03508 & \textbf{0.02824} & 0.71082 & 0.15410 & 0.25833 \\
& BuSNN~\cite{zhang2026burst} & 0.01151 & 0.07751 & 0.06301 & 0.95553 & 0.18205 & 0.30903 \\
\midrule
NSD & ANN~\cite{ResNet18} & 0.02331 & 0.66545 & 0.73041 & 0.99581 & 0.22095 & 0.20642 \\
& LIF~\cite{lif} & \textbf{0.00056} & \textbf{0.02169} & \textbf{0.01231} & \textbf{0.99586} & 0.20999 & 0.15061 \\
& PSN~\cite{psn} & 0.00195 & 0.06067 & 0.05984 & 0.99570 & 0.22460 & 0.23207 \\
& MPSN~\cite{mpsn} & 0.00221 & 0.02750 & 0.02820 & 0.96882 & 0.18208 & 0.19292 \\
& BuSNN~\cite{zhang2026burst} & 0.00190 & 0.05692 & 0.05853 & 0.99570 & \textbf{0.23630} & \textbf{0.23390} \\
\bottomrule
\end{tabular}
\end{table}

\subsection{{fMRI Semantic Classification}}

To examine whether the fMRI-decoded features support semantic interpretation, we compared category predictions obtained from fMRI-decoded features with those obtained from model-derived features. For each test stimulus, both features were passed through the classifier head of the corresponding ANN or SNN backbone. We evaluated semantic decoding using Top-1 agreement.
These metrics measure whether the decoded representation preserves the category-level prediction and the candidate-level semantic structure of the model-derived features.

Top-1 agreement measures whether the most confident category predicted from the fMRI-decoded feature matches that predicted from the corresponding model-derived feature. As shown in Table~\ref{tab:semantic_main}, SNN-based representations consistently improve Top-1 agreement over the ANN baseline. On GoD, the best Top-1 agreement increases from 0.1800 for ANN to 0.4400 with the LIF-based SNN. On NSD, the best score increases from 0.2342 for ANN to 0.4206 with PSN. These results indicate that, under the same fMRI-to-feature decoder, SNN-derived representations preserve category-level semantic information more effectively than ANN features.


\begin{table}[!t]
\centering
\caption{Top-1 semantic decoding performance on the validation and test sets. Higher is better.}
\label{tab:semantic_main}
\setlength{\tabcolsep}{6pt}
\begin{tabular}{llccc}
\toprule
Dataset & Model & Train & Val & Test \\
\midrule
GoD & ANN~\cite{ResNet18} & 0.94167 & 0.23333 & 0.18000 \\
& LIF~\cite{lif} & 0.95740 & 0.47500 & \textbf{0.44000} \\
& PSN~\cite{psn} & \textbf{0.97314} & \textbf{0.54167} & 0.40000 \\
& MPSN~\cite{mpsn} & 0.94815 & 0.32500 & 0.38000 \\
& BuSNN~\cite{zhang2026burst} & 0.95740 & 0.35833 & 0.32000 \\
\midrule
NSD & ANN~\cite{ResNet18} & \textbf{1.00000} & 0.40000 & 0.23422 \\
& LIF~\cite{lif} & {0.10000} & \textbf{0.60000} & 0.32994 \\
& PSN~\cite{psn} & \textbf{1.00000} & \textbf{0.60000} & \textbf{0.42057} \\
& MPSN~\cite{mpsn} & 0.96296 & 0.53333 & 0.28819 \\
& BuSNN~\cite{zhang2026burst} & {0.10000} & 0.40000 & 0.35743 \\
\bottomrule
\end{tabular}
\end{table}

\begin{figure}[!t]
\centering
\includegraphics[width=\columnwidth]{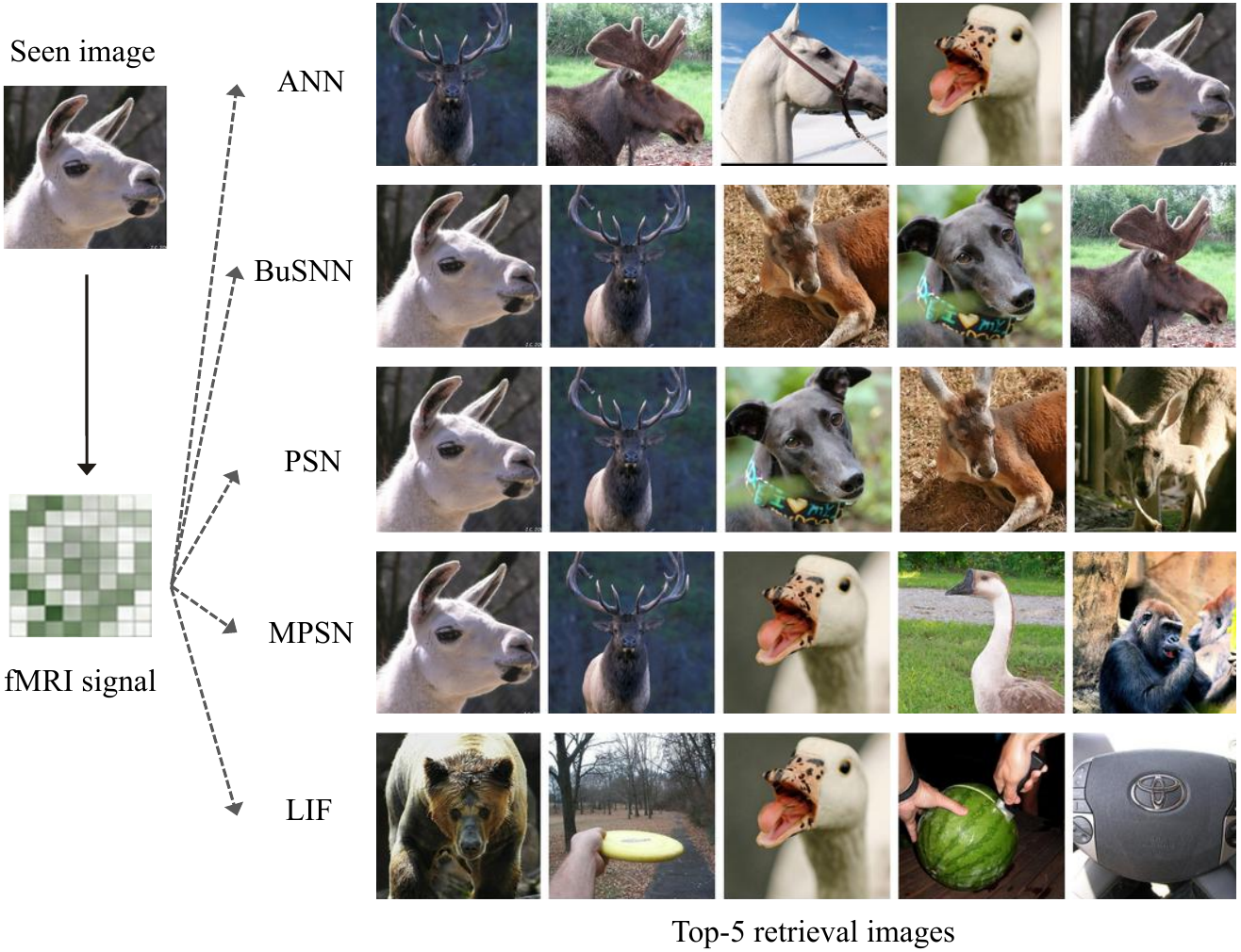}
\caption{{Qualitative comparison of ANN- and SNN-based fMRI-to-image retrieval on the GoD dataset.}}
\label{fig:retrieval}
\end{figure}

\subsection{{fMRI-to-image Retrieval}}

We evaluated fMRI-to-image retrieval on the GoD dataset. For each test sample, the fMRI response elicited by the viewed stimulus was mapped into the corresponding ANN or SNN {visual features}. The decoded feature was then used as a query to rank candidate stimulus images according to {cosine similarity in the corresponding feature space}.

Retrieval performance was measured by Acc@$K$ with $K=1,3,5$. A retrieval was counted as correct when the ground-truth stimulus image appeared among the top-$K$ retrieved candidates.
Table~\ref{tab:retrieval} reports the retrieval results. Compared with the ANN baseline, most SNN {visual features} achieve higher retrieval accuracy. The PSN model obtains the best Acc@1 and Acc@3, improving Acc@1 from 0.44 to 0.58 and Acc@3 from 0.70 to 0.88. For Acc@5, PSN and MPSN both reach the highest accuracy of 0.94, compared with 0.82 for the ANN baseline. BuSNN also improves over the ANN baseline across all ranks, while LIF performs slightly lower than ANN in this retrieval setting.
Fig.~\ref{fig:retrieval} provides qualitative retrieval examples on the GoD dataset. We show the ground-truth stimulus image and the top retrieved candidates obtained from ANN- and SNN-based decoding. These examples provide an interpretable comparison of how closely the retrieved candidates match the viewed stimulus under different {visual features}.

\subsection{{fMRI-to-image Reconstruction}}

We further evaluated whether the decoded semantic information can support image reconstruction. 
Following the semantic-guided reconstruction strategy used in fMRI decoding~\cite{ferrante2024retrieving}, the decoded feature was first converted into a semantic prediction. 
The predicted category was then used as a text prompt to condition a pretrained text-to-image diffusion model~\cite{ho2020denoising}. 
Visual feature extraction was performed separately from the reconstruction stage, and the diffusion model generated visualization results. The generator was kept fixed, so the reconstruction examples were used only as qualitative visualizations of the semantic information preserved in the decoded features.


To quantify reconstruction quality, we define a best-of-five semantic reconstruction score, denoted as $\mathrm{Bo5\text{-}SRC}$. 
For each sample, five candidate images are generated and the one with the highest feature similarity to the ground-truth stimulus is selected. 
The final score jointly considers three aspects: feature similarity, retrieval-rank consistency, and semantic-category consistency. Specifically,
\begin{equation}
\mathrm{Bo5\text{-}SRC}
=
\frac{1}{|\mathcal{D}_{\mathrm{te}}|}
\sum_{i}
\left(
0.5 C_i + 0.3 R_i + 0.2 U_i
\right),
\end{equation}
where $C_i$ denotes the normalized cosine similarity in feature space, $R_i$ measures the relative ranking of the ground-truth image among all test stimuli, and $U_i$ denotes the Wu--Palmer similarity between predicted and ground-truth categories.

Table~\ref{tab:reconstruction} reports the reconstruction results on GoD and NSD. 
On GoD, PSN achieves the best score, improving $\mathrm{Bo5\text{-}SRC}$ from 0.7747 for the ANN baseline to 0.8087. 
BuSNN also improves over the ANN baseline, reaching 0.8059, while LIF obtains a smaller gain with 0.7936. 
On NSD, PSN again achieves the best score, increasing $\mathrm{Bo5\text{-}SRC}$ from 0.7402 to 0.7553, followed by BuSNN with 0.7524. 
MPSN and LIF do not improve over the ANN baseline on this metric. 
These results suggest that SNN-derived features yield reconstruction benefits.

\begin{figure}[!t]
\centering
\includegraphics[width=0.9\columnwidth]{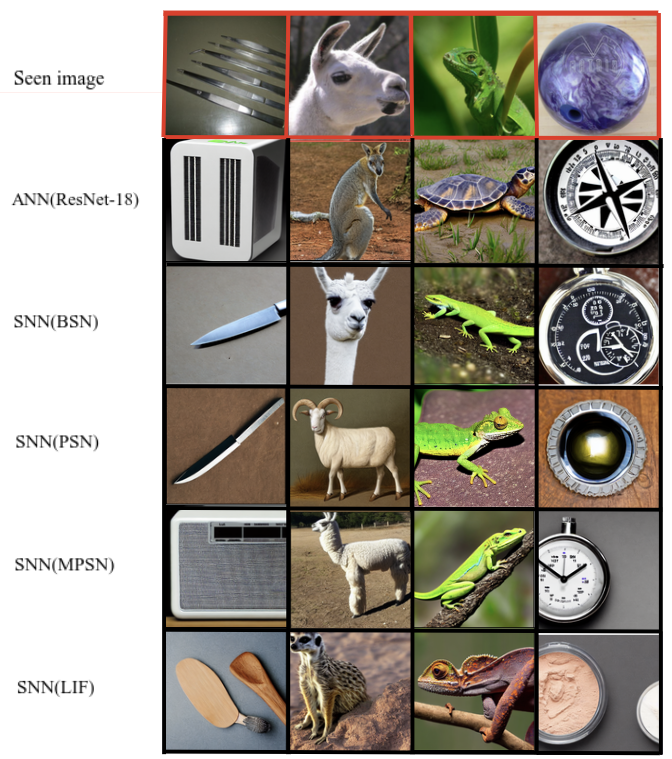}
\caption{Qualitative comparison of ANN- and SNN-based fMRI-to-image reconstruction on the GoD dataset.}
\label{fig:reconstruction}
\end{figure}

Fig.~\ref{fig:reconstruction} shows qualitative reconstruction examples on the GoD dataset. 
For each fMRI query, we compare the ground-truth stimulus with images generated from ANN- and SNN-derived semantic predictions. 
The examples provide a visual comparison of how different {visual features} affect the semantic content of the reconstructed images.

\begin{table}[!t]
\centering
\caption{fMRI-to-image retrieval accuracy on GoD. Higher is better.}
\label{tab:retrieval}
\setlength{\tabcolsep}{5pt}
\begin{tabular}{lccccc}
\toprule
 & ANN~\cite{ResNet18} & LIF~\cite{lif} & PSN~\cite{psn} & MPSN~\cite{mpsn} & BuSNN~\cite{zhang2026burst} \\
\midrule
Acc@1 & 0.44 & 0.40 & \textbf{0.58} & 0.52 & 0.56 \\
Acc@3 & 0.70 & 0.66 & \textbf{0.88} & 0.82 & 0.84 \\
Acc@5 & 0.82 & 0.80 & \textbf{0.94} & \textbf{0.94} & 0.92 \\
\bottomrule
\end{tabular}
\end{table}

\begin{table}[!t]
\centering
\caption{Semantic-guided fMRI-to-image reconstruction performance on GoD and NSD measured by $\mathrm{Bo5\text{-}SRC}$. Higher is better.}
\label{tab:reconstruction}
\setlength{\tabcolsep}{5pt}
\begin{tabular}{lccccc}
\toprule
Dataset & ANN~\cite{ResNet18} & LIF~\cite{lif} & PSN~\cite{psn} & MPSN~\cite{mpsn} & BuSNN~\cite{zhang2026burst} \\
\midrule
GoD & 0.7747 & 0.7936 & \textbf{0.8087} & 0.7571 & 0.8059 \\
NSD & 0.7402 & 0.6993 & \textbf{0.7553} & 0.7215 & 0.7524 \\
\bottomrule
\end{tabular}
\end{table}

\subsection{Ablation Analysis}

We conducted ablation analyses to further examine which aspects of the SNN representation contribute to the observed decoding gains. 
Specifically, we analyzed two factors: the effect of spiking neuron dynamics and the effect of time steps. 
The first analysis compares different SNN variants under the same fMRI decoding framework, while the second fixes the neuron model to LIF and varies the number of time steps. {This design separates two questions that are often conflated in SNN applications: whether the form of spike generation matters, and whether the amount of temporal accumulation matters once the neuron type is fixed.}

\subsubsection{Effect of spiking neuron dynamics}

We first examined the influence of different spiking neuron dynamics by comparing LIF~\cite{lif}, PSN~\cite{psn}, MPSN~\cite{mpsn}, and BuSNN~\cite{zhang2026burst} models. 
The results in the preceding sections show that different spiking dynamics lead to different decoding behaviors. 

As shown in Table~\ref{tab:combined_alignment}, LIF and MPSN achieve the lowest GoD test MSE of 0.0282, while LIF obtains the lowest NSD test MSE of 0.0123. These results suggest that membrane-based integration and memory-supported propagation produce features that are more readily predicted from fMRI responses under a linear decoder.
The ranking changes when semantic-level metrics are considered. Table~\ref{tab:semantic_main} shows that LIF achieves the best GoD Top-1 agreement of 0.4400, whereas PSN obtains the best NSD Top-1 agreement of 0.4206. PSN also achieves the best GoD retrieval accuracy, with Acc@1/3/5 values of 0.58/0.88/0.94 (Tables~\ref{tab:retrieval}), and obtains the highest semantic-guided reconstruction score on both GoD and NSD, as shown in (Tables~\ref{tab:reconstruction}). BuSNN also performs competitively in retrieval and reconstruction, indicating that burst-like spike coding can preserve fMRI-recoverable semantic cues.

Overall, these results indicate that the advantage of SNN representations is not attributable to a single neuron implementation. 
Different spiking dynamics emphasize different aspects of the representation: membrane-based and memory-based dynamics are more favorable for low-error semantic-level alignment, whereas parallel and burst-like spike formulations can be more effective for preserving semantic information used in retrieval and reconstruction.

\subsubsection{Effect of time steps}

We next examined the role of temporal processing by fixing the neuron model to LIF and varying the number of time steps. 
The default LIF model used in the main experiments corresponds to $T=4$, and we additionally evaluated shorter and longer time steps with $T=2$ and $T=8$. 
This analysis isolates the effect of temporal accumulation from the effect of changing the neuron model.

\begin{figure}[!t]
\centering
\includegraphics[width=0.49\columnwidth, keepaspectratio=false]{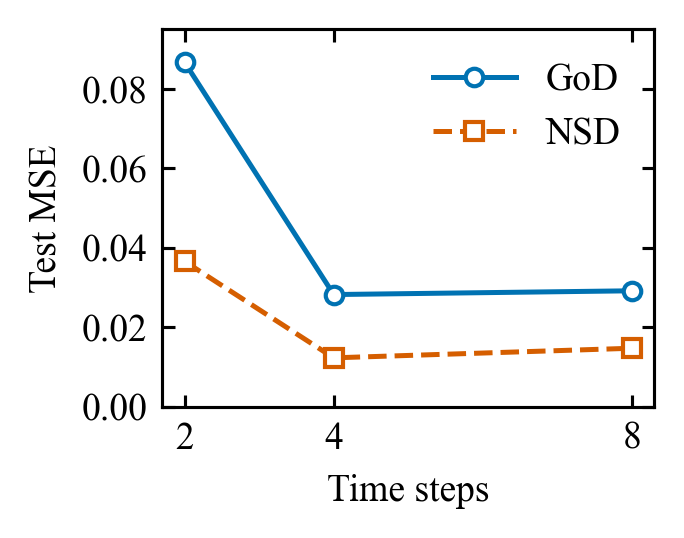}
\hfill
\includegraphics[width=0.49\columnwidth, keepaspectratio=false]{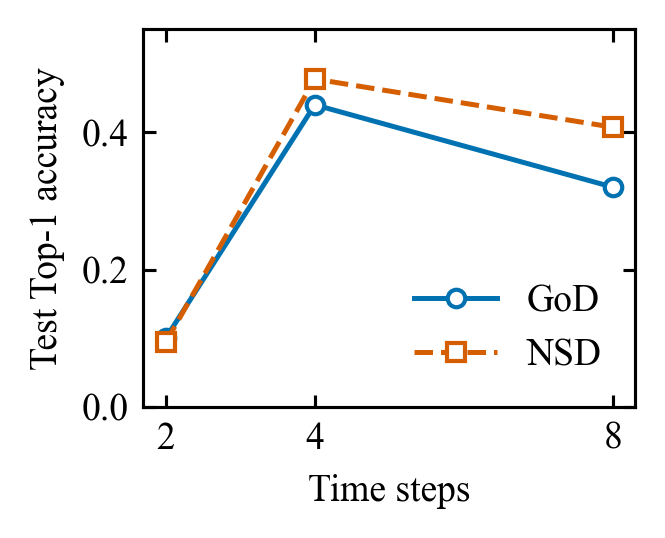}
\caption{Effect of time steps under the LIF neuron model on test-set semantic-level representation alignment error and Top-1 semantic decoding accuracy.}
\label{fig:ablation_time}
\end{figure}

Fig.~\ref{fig:ablation_time} shows the effect of time steps on both semantic-level representation alignment error (MSE, left) and Top-1 semantic agreement (right). On both GoD and NSD datasets, the default setting $T=4$ achieves the lowest test MSE and highest semantic agreement. Results of $T=2$ performs substantially worse due to insufficient temporal spike accumulation. Although $T=8$ improves over $T=2$, it does not surpass $T=4$. These results indicate that an appropriate time step is crucial for preserving semantic information in SNN representations, as too few steps limit decodability while overly long sequences may introduce redundant or less stable responses.

Overall, the ablation results support two conclusions. 
First, spiking neuron dynamics influence the type of information emphasized in the model-derived features. 
Second, the number of time steps affects both feature predictability and semantic decoding quality. 
The best performance is obtained when temporal dynamics are present but properly controlled, supporting the use of temporally structured SNN representations for fMRI-based visual decoding.

\section{Discussion}
\label{sec:discussion}

This study suggests that the {visual features} used for fMRI decoding should be treated as a scientific hypothesis, rather than merely as an engineering choice. 
Under the same fMRI-to-feature decoder, changing only the {predicted visual features} affected both feature predictability and downstream semantic fidelity. 
Across GoD and NSD, SNN-derived {visual features} were more predictable from fMRI responses and produced stronger semantic agreement than the ANN baseline built on the same architectural family.
The consistency across multiple spiking neuron models further suggests that the observed advantage is not tied to a single SNN implementation.

{From a neuroimaging perspective, the consistent advantage of SNN-derived features suggests that sparse and temporally structured visual representations may better match the coarse and temporally integrated nature of BOLD measurements than dense ANN activations.} Different spiking dynamics further emphasize different aspects of the decoded representation: LIF tends to improve feature predictability, whereas PSN is more competitive in semantic retrieval and reconstruction.
{We do not interpret this result as evidence that fMRI directly measures biological spikes. Instead, we interpret it as evidence that spiking representations can provide a more decodable intermediate {representation}.} This interpretation is consistent with evidence that BOLD signals reflect synaptic and local population processing more closely than spiking output alone~\cite{logothetis2001neuro,panzeri2015neural}. {It is also compatible with the stronger voxel-level PCC distributions in Fig.~\ref{fig:pcc}, which indicate a closer correspondence between SNN features and ROI-level responses.}

{This work has several limitations.}
{First, we use the same fMRI-to-feature decoder across all {visual features} to ensure a controlled comparison of representation differences. This design helps isolate the effect of the target feature space, although more expressive subject-specific decoders could further improve absolute decoding performance.}
{Second, our comparison focuses on matched ANN and SNN backbones, which allows us to control architectural differences between the two model families. Future studies could extend this analysis to additional ANN architectures, such as vision transformers~\cite{dosovitskiyimage} or multimodal models~\cite{clip}, to provide a broader characterization of SNN-derived {visual features} in relation to diverse ANN representations.}
{Furthermore, the reconstruction analysis is based on semantic-guided generation rather than direct pixel-level reconstruction from fMRI activity.}
{Future work may combine SNN-derived representations with stronger generative models, broader visual-model baselines, subject-level statistical testing, and human perceptual evaluation to better characterize reconstruction quality and robustness.}

\section{Conclusion}
\label{sec:conclusion}

In this work, we evaluated whether the target {visual representation} itself can improve fMRI-based visual decoding. 
Across multiple fMRI benchmarks, SNN-derived visual features provided stronger decoding results than an ANN baseline from the same architectural family. 
The improvements appeared in voxel-level alignment, semantic-level alignment, and semantic decoding. 
{Ablation analyses further show that both spiking neural dynamics and time steps contribute to this advantage. 
These results indicate that temporally structured spike-based representations provide a more favorable decoding target than static dense activations under the same fMRI-to-feature decoder.} 
{Overall, our findings support a broader view that representation design is an important part of the neuroimaging decoding model. 
They also demonstrate that SNN-derived representations provide effective {visual features} for fMRI-based visual decoding.}



\section*{Data Availability}

The code will be released upon publication.

\bibliographystyle{IEEEtran}
\bibliography{ref}

\end{document}